%% file: main.tex
\PassOptionsToPackage{table}{xcolor}

\documentclass[10pt,twocolumn,letterpaper]{article}

\usepackage{cvpr}              


\definecolor{cvprblue}{rgb}{0.21,0.49,0.74}
\usepackage[pagebackref,breaklinks,colorlinks,allcolors=cvprblue]{hyperref}
\usepackage{url}            
\usepackage{booktabs}       
\usepackage{amsfonts}       
\usepackage{nicefrac}       
\usepackage{microtype}      
\usepackage{xcolor}
\usepackage{graphicx}
\usepackage{caption}
\usepackage{subcaption}
\usepackage{amsmath}
\usepackage{svg}
\usepackage{pifont}
\usepackage{algorithm}
\usepackage{algpseudocode}
\usepackage{amsmath}
\usepackage{amssymb}
\usepackage{fullpage}
\usepackage{amsfonts}
\usepackage{multirow}
\usepackage{float}

\def\paperID{830} 
\def\confName{CVPR}
\def\confYear{2025}

\title{RadSAM: Segmenting 3D radiological images with a 2D promptable model}

\author{
    Julien Khlaut$^{1,2}$ \quad
    Elodie Ferreres$^{1}$ \quad
    Daniel Tordjman$^{1}$ \quad
    Hélène Philippe$^{2}$ \quad
    Tom Boeken$^{2}$ \\
    Pierre Manceron$^{1}$ \quad
    Corentin Dancette$^{1}$ \\
    $^1$Raidium, France \\
    $^2$Université de Paris Cité, AP-HP, Hôpital Européen Georges Pompidou, \\
    Department of Vascular and Oncological Interventional Radiology, \\
    HEKA INRIA, INSERM PARCC U 970, Paris, France
}

\newcommand\smallpm[1]{\footnotesize{$\pm #1$}}

\begin{document}
\maketitle
\begin{abstract}
\input{sec/0_abstract} 
\end{abstract}
\input{sec/1_intro}
\input{sec/2_method}

\input{sec/3_results}

\clearpage

{
    \small
    \bibliographystyle{ieeenat_fullname}
    \bibliography{references, others}
}

\input{sec/X_suppl}


\end{document}

%% file: sec/0_abstract.tex
Medical image segmentation is a crucial and time-consuming task in clinical care, where mask precision is extremely important. The Segment Anything Model (SAM) offers a promising approach, as it provides an interactive interface based on visual prompting and edition to refine an initial segmentation. This model has strong generalization capabilities, does not rely on predefined classes, and adapts to diverse objects; however, it is pre-trained on natural images and lacks the ability to process medical data effectively. In addition, this model is built for 2D images, whereas a whole medical domain is based on 3D images, such as CT and MRI. Recent adaptations of SAM for medical imaging are based on 2D models, thus requiring one prompt per slice to segment 3D objects, making the segmentation process tedious. They also lack important features such as editing. 
To bridge this gap, we propose RadSAM, a novel method for segmenting 3D objects with a 2D model from a single prompt. In practice, we train a 2D model using noisy masks as initial prompts, in addition to bounding boxes and points. We then use this novel prompt type with an iterative inference pipeline to reconstruct the 3D mask slice-by-slice. We introduce a benchmark to evaluate the model's ability to segment 3D objects in CT images from a single prompt and evaluate the models' out-of-domain transfer and edition capabilities. We demonstrate the effectiveness of our approach against state-of-the-art models on this benchmark using the AMOS abdominal organ segmentation dataset.

%% file: sec/1_intro.tex
\section{Introduction}
\label{sec:intro}

Image segmentation is an essential task for medical imaging~\cite{pham2000current}: it allows specialists to compute multiple metrics regarding anatomical and pathological objects useful for clinical care. 
The manual segmentation process is long, error-prone, subjective, and led to the adoption of simplified criteria such as RECIST~\cite{Eisenhauer2009Newrecist} to avoid segmenting whole 3D structures.
Models that streamline time-consuming tasks, such as estimating the volume of a liver prior to treatment or quantifying a patient's cancer burden through extensive tumor segmentation, help physicians in their everyday decision-making and reduce inconsistencies between multiple annotators.

Deep-learning-based methods for organs and tumors have been studied extensively; most of them are trained on specific datasets using U-net architectures \cite{isensee2021nnu,hatamizadeh2022unetr,Milletar2016VNetFC,Zhou2018UNetAN}. However, standard semantic segmentation models are subject to limitations in their clinical usage. First, once the model outputs a segmentation map for a given image, it has no mechanism to correct it. In case of an error, the user has to modify it manually, which is time-consuming. 

Recently, the Segment Anything Model (SAM)~\cite{kirillov_segment_2023} was introduced and proposed a prompted segmentation framework, allowing users to interactively guide the model with a spatial prompt to obtain an initial segmentation. Users can then provide correction points, helping the model refine the mask. 
Moreover, being trained on very large and diverse datasets, SAM has the ability to segment objects that are not necessarily present in the initial dataset. It is, therefore, considered a foundation model for segmentation.
This framework is promising for medical image segmentation, as it allows the precise segmentation of diverse structures, with more consistency and speed than manual segmentation. 
However, the original Segment Anything model (SAM) was shown to be unreliable for medical data~\cite{mazurowski_segment_2023,zhang_segment_2024}. Thus, Ma et al. proposed MedSAM~\cite{ma_segment_2024}, a fine-tuned SAM for medical segmentation. SAM-Med 2D~\cite{cheng_sam-med2d_2023},  Medical SAM Adapter~\cite{wu2023medical} and SA-Med2D~\cite{ye2023samed2D} feature similar approaches. Yet, they lack important SAM features, as shown in Table~\ref{tab:model_comparaison}. For instance, MedSAM only supports bounding-box prompts and does not implement SAM's editing mode.
More importantly, in radiology, high-quality imaging is produced in 3D, with Magnetic resonance imaging (MRI) and Computed Tomography (CT), which requires models to segment 3D structures.
Previous SAM-based models only propose a 2D segmentation method, requiring one prompt for each 2D slice to segment full 3D objects, with potentially noncoherent segmentation for consecutive slices.
Some work, such as SAM-Med-3D~\cite{wang2023sammed3d}, propose a model that directly segments 3D medical images from the full volume. However, such a 3D model has high memory GPU constraints, making it hard to train and deploy. Indeed, during training, the model needs to have access to the whole volume, which uses more memory than the 2D-based model.  It also outputs a lower resolution mask, 128x128, compared to the standard 256x256 outputted by SAM.
We chose to focus on 2D models because they are: easy to implement and train, efficient, and produce higher-resolution masks.

In the present work, we introduce RadSAM (for Radiological SAM). This promptable segmentation model can segment 3D structures in CT images from a single prompt, using the memory footprint of a 2D model.
Our method is simple yet effective: in addition to points and box prompts, we train a 2D segmentation model with a mask prompt as the first input. The model learns to reconstruct the ground-truth mask from a degraded input.
We then introduce an iterative inference method to leverage the novel mask prompt and forward instructions from one slice to the next with minimal information loss.
This allows us to segment 3D radiological objects from a single prompt and to add 3D editing capabilities.
We benchmark our promptable segmentation model on 3D imaging, using AMOS~\cite{ji_amos_2022} and TotalSegmentator~\cite{wasserthal_totalsegmentator_2023}, two CT organ segmentation datasets. We evaluate its capacity to segment from multiple prompt types, transferability to other datasets, and edition performances.
We will release our model weights upon acceptance of the paper.

\begin{table}[h]
    \centering
    \small
    \begin{tabular}{ccccc}
    \toprule
    Model    &  Domain &  3D          &    Prompt       &    Correction   \\ \midrule
    SAM      &  Natural      &  \ding{55}   &   Bbox / Point    &    \ding{51}   \\
    MedSAM   &  Medical      &  \ding{55}   &   Bbox    &    \ding{55}   \\
    RadSAM   &  Medical      &  \ding{51}   &   Bbox / Point\textsuperscript{\textdagger} & \ding{51}   \\
    \bottomrule
    \end{tabular}
    \caption{Different settings and characteristics of the SAM-like models. \textsuperscript{\textdagger}RadSAM is also trained with mask prompts, but they are not used as the first prompt during inference.}
    \label{tab:model_comparaison}
\end{table}

%% file: sec/2_method.tex
\section{Benchmark}
\subsection{Datasets}
To produce our benchmark, we consider only models trained on the AMOS dataset~\cite{ji_amos_2022}. It comprises a large and diverse collection of clinical CT scans for abdominal organ segmentation and some MRIs. We only use the CT part containing 500 CT scans and voxel-level annotations of 15 abdominal organs. In addition to AMOS, we use another dataset for additional experiments: TotalSegmentator (TS)~\cite{wasserthal_totalsegmentator_2023}: This dataset consists of 1204 CT scans with detailed annotations of 104 anatomical structures (27 organs, 59 bones, 10 muscles, and 8 vessels).

\subsection{Input prompts}
\begin{figure}[h]
    \centering
    \includegraphics[width=\linewidth]{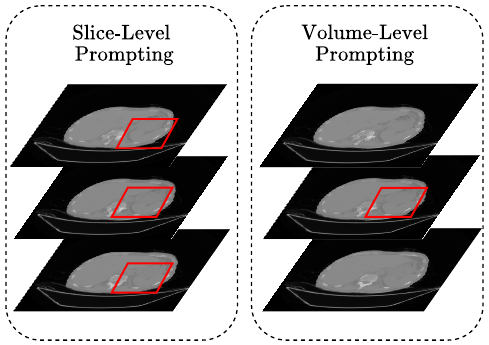}
    \caption{Prompting strategies. Slice-level prompting consists of feeding the model with one prompt for each slice. Volume-level prompting consists of a single 2D prompt (point or bbox) and annotations for the top and bottom slices.}
    \label{benchmark}
\end{figure}

The benchmark evaluates the model's response to different prompting strategies, as detailed in Figure \ref{benchmark}. We consider 2 approaches. The first one is slice-level prompting, which consists of predicting a mask for each slice of interest by giving a prompt on each slice. The second one is volume-level prompting; this consists of a unique 2D prompt on one slice and 2 boundary annotations indicating the maximum and minimum slices.
We also study the models' ability to respond to edition prompts. For the 2D slice-level prompting, we add a point to each of the slices, whereas for the volume-level prompting with boundaries, we add a point for the whole volume.

\section{RadSam Training}
\label{sec:method}

In this section, we detail the RadSAM architecture and training procedure. We start by leveraging SAM's architecture and adapting it to segment 2D slices of medical images.

\subsection{Architecture}

\begin{figure}[h]
    \centering
    \includegraphics[width=\linewidth]{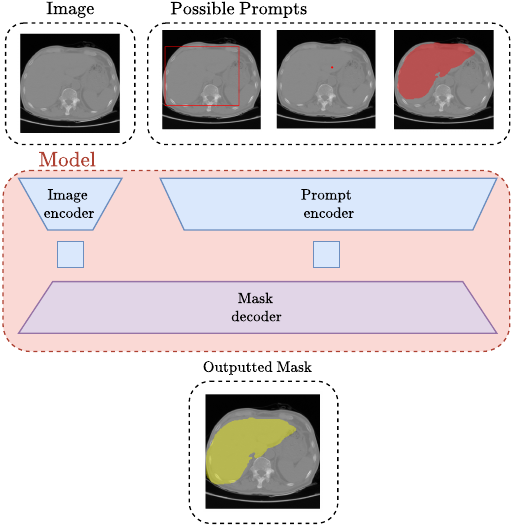}
    \caption{RadSAM Architecture. The available prompts are masks, boxes, and points.}
    \label{fig:raps-training}
\end{figure}

We use the SAM~\cite{kirillov_segment_2023} architecture and weights as a starting point to train RadSAM. 
The model, as shown in Figure~\ref{fig:raps-training}, takes as input a 2D image $v \in \mathbb{R}^{3\times H \times W}$  and a combination of one or multiple prompts. Possible prompts include a bounding-box $p_\mathrm{box} \in \mathbb{R}^{4} $, one or multiple points $p_\mathrm{point}\in \mathbb{R}^{N \times 2}$ along with their positive/negative labels $p_l\in[0, 1]^{N} $, or a mask $p_m \in \mathbb{R}^{H' \times W'}$, and returns one or multiple masks $m \in R^{H\times W}$, where $H'$ and $W'$ are the low-resolution dimensions of SAM output ($224 \times 224$). Some prompts are optional depending on the training mode, which will be detailed in the next sections. The image encoder (a ViT~\cite{Dosovitskiy2020AnII}) processes the image to obtain image features. The prompt encoder module processes the prompts, and the mask decoder uses the image and the prompt features to return one or multiple logits masks $m$. We always generate 4 masks: one primary and three secondary masks, used for oracle prediction. For all evaluations, we use the primary mask unless specified otherwise.
We keep the SAM architecture unchanged but modify the training procedure to allow for new prompt combinations. The architecture is shown in Figure~\ref{fig:raps-training}.

\subsection{Preprocessing and data curation}
The images used in this study are CT scans. They are 3D arrays with values from around -1000 to +1000 Hounsfield Units (HU). In contrast, pixel intensity values within each image are limited to a relatively narrow range of 0 to 255. We apply thresholding to all slices within the dataset, restricting intensity values to the range of -500 to 1000 HU. Subsequently, these values are rescaled to $[0, 1]$. To filter out images with sparse or noisy content, we discard slices that contain fewer than 15 pixels. Finally, each image is resized and padded to a uniform size of $1024 \times 1024$ and copied across the three RGB channels to fit into the SAM image encoder.
We then create training samples of all potential triplets ($v$, $p$, $g$): the input slice, the prompt (point, bounding box, or mask), and the ground truth mask.

\begin{figure}[h]
    \centering
    \includegraphics[width=\linewidth]{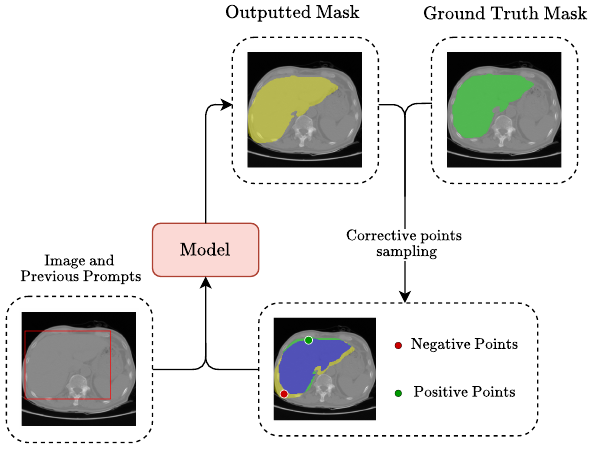}
    \caption{RadSAM edition training pipeline. The mask decoder outputs a mask refined through editing points using the ground truth mask. The outputted mask is represented in yellow, the prompt in red, the ground truth mask in green, and the intersection between the ground truth and the outputted mask is in blue.}
    \label{fig:raps-edition-training}
\end{figure}

\subsection{Prompts generation}
\begin{figure}
     \centering
     \includegraphics[width=\linewidth]{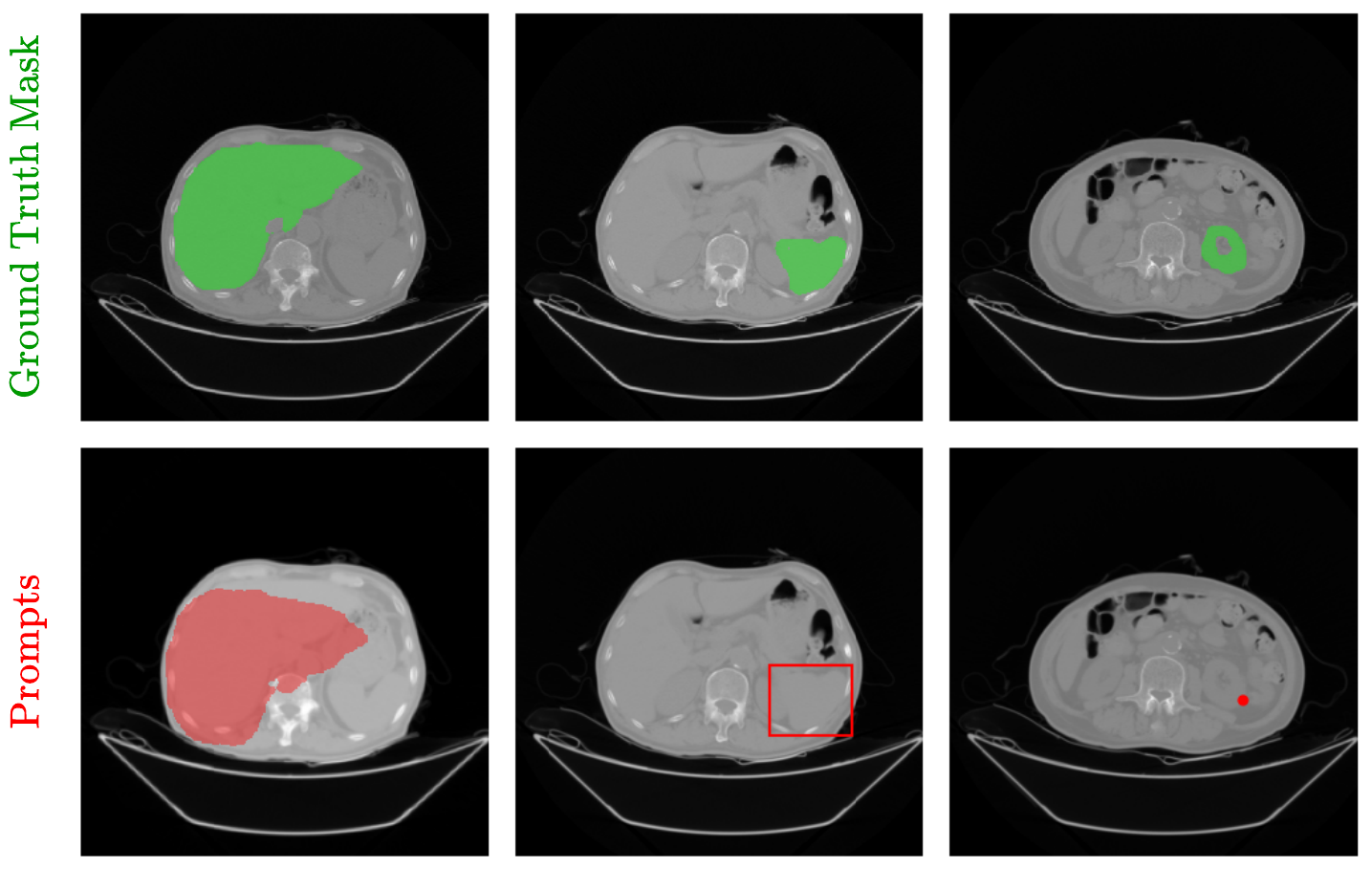}
    \caption{Examples of each prompt: mask, bounding box, and point.}
        \label{fig:prompts}
\end{figure}

We generate three prompts for each object to train and evaluate our model: point, bounding box, and mask. The prompts are randomly created from the ground truth mask.
\begin{itemize}
    \item Point: a random location is selected from within the ground truth mask, avoiding 2 pixels along its contour.
    \item Box: we generate a bounding box that encloses the mask. The box dimensions are randomly perturbed between -5 to +20 pixels on each side.
    \item Mask: We generate a noisy version of the ground-truth mask with random transforms (rotation, scaling, translation, erosion, and dilation).
    This novel prompt type enables iterative generation (Sec.~\ref{sec:method-ar}) or manually drawing a mask.
    \item{Correction points}: We train RadSAM with iterative correction points. Each point is sampled uniformly in the error mask (false negative and false positive). The point is then assigned its label (respectively positive or negative). This pipeline is displayed in Figure~\ref{fig:raps-edition-training}.
\end{itemize}
Figure~\ref{fig:prompts} shows an example for each prompt.
Introducing masks as input is a new feature, which poses some challenges. SAM was trained on mask logits values rather than binary masks, but our input masks are binary. To unify the input prompts, we encode all masks (first input or edition mask) with binary values instead of logits.



\subsection{Training procedure}

The training objective we use is the sum between the dice loss~\cite{milletari2016vdice}, commonly used for medical imaging, and the binary cross-entropy loss.
The dice loss is defined as such:
\vspace{-0.1cm}

$$\mathcal{L}_{d} = 1 - 2 \frac{\sum_i^B m_i g_i}{\sum_i^B m_i^2 + \sum_i^B g_i^2} $$

Where $B$ is the batch size, $g_i$ is the ground truth mask for the task $i$, and $m_i$ is the mask outputted by the model.
The binary cross entropy is defined as such:

$$ \mathcal{L}_{\mathrm{BCE}} = - (\frac{1}{B} \sum_i^B g_i \log m_i + (1 - g_i) \log (1 - m_i)) $$

The final loss is $\mathcal{L} = \mathcal{L}_{ds}  + \mathcal{L}_{\mathrm{BCE}}$
. During the training, the model generates 4 masks for each task. The loss is always applied to the first (primary mask) and to the best of the three other masks.
A similar procedure is done for editing steps, and the final loss is averaged over the steps and the batch. The pipeline of the edition training is described in Figure \ref{fig:raps-edition-training}.

Training is performed using 8 nodes, each equipped with 4 V100 GPUs, over a period of 40 hours, which corresponds to approximately 4 epochs of training.

To enhance the model's generalization, we augment the data with random translations, rotations, shear, zoom, and Gaussian noise.
In addition, we also randomly vary the windowing used to clip the HU values. 


\section{RadSAM with Volume-Level Prompting}
\label{sec:method-ar}
As mentioned in Section~\ref{sec:intro}, segmenting 3D objects using one slice per prompt is very time-consuming for users. We introduce an iterative inference method to segment 3D objects with a single prompt. This method is compatible with every 2D promptable model but can leverage RadSAM's novel mask input prompt for better results.

\subsection{Iterative Segmentation}
\label{sec:method-ar:iterative}
\begin{figure}[h]
    \centering
    \includegraphics[width=\linewidth]{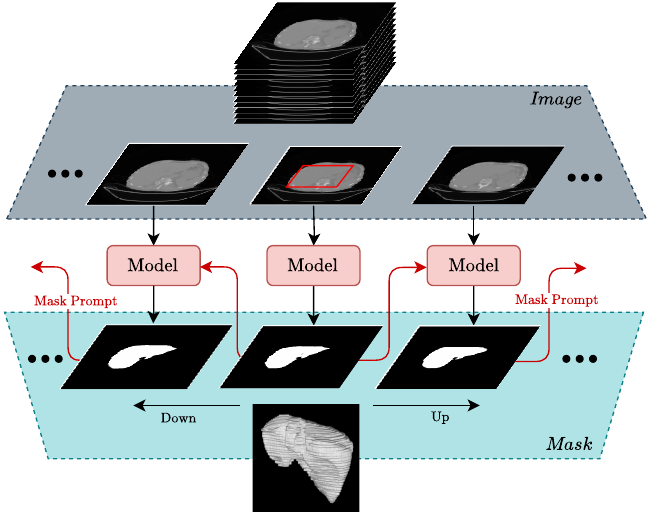}
    \caption{The iterative segmentation pipeline. The model is applied to each slice independently, starting from the slice where the first prompt is. Then, the mask output is passed as a prompt to segment the next slice.}
    \label{fig:auto-regressive}
\end{figure}

We propose an inference pipeline where the user provides a single 2D prompt with top and bottom slice boundaries. We display the inference pipeline in Figure~\ref{fig:auto-regressive}. 
The user provides an initial prompt: a point, a bounding box, or a mask on a single slice $i$. The segmentation model returns a mask $m_i$ for this slice.
We then create a new prompt for the slices $i+1$ and $i-1$ based on $m_i$. For RadSAM, we simply use the mask $m_i$ as the new prompt. For models that do not support the initial mask prompting, such as SAM or MedSAM, we generate a bounding box around this mask. 
Additionally, we use two boundary annotations (top and bottom slices of the object) to stop the propagation.

\begin{table*}[t]
\centering
\begin{tabular}{lcrcccccc}
    \toprule
    & \multicolumn{4}{c}{\small Volume-level prompt} &  \multicolumn{4}{c}{\small Slice-level prompt}  \\
    \cmidrule(r){2-5} \cmidrule(l){6-9}
    \small{Model} & \footnotesize{\#Prompts} & \footnotesize{\#Edits} & {\small Bbox} & \small Point & 
    \footnotesize{\#Prompts} & \footnotesize{\#Edits} &{\small Bbox} & \small Point \\
    \midrule
    nnU-Net~\textsuperscript{\textdagger} & $\varnothing$ & $\varnothing$ & \multicolumn{2}{c}{ \cellcolor[HTML]{e6e6e6}  $88.87$}  & $\varnothing$ & $\varnothing$ & \multicolumn{2}{c}{ \cellcolor[HTML]{e6e6e6}  $88.87$}\\ \midrule
     \multirow{2}{*} {SAM-Med3D} & $1$ & - & - & $79.94$  \\
     & $1$ & 9 & - & $83.99$  \\ \midrule
    \multirow{2}{*}{SAM} & $3$ & --  & $50.93$ \smallpm{1.40} & $38.78$ \smallpm{1.73} & $n$  & -- & $70.20$ \smallpm{0.90} & $33.07$ \smallpm{1.60} \\ 
    & $3$ & 20  & $60.60$ \smallpm{1.36} & $47.16$ \smallpm{1.74} & $n$  & $10n$ & $85.43$ \smallpm{0.35} & $84.31$ \smallpm{0.45} \\ \midrule
    MedSAM& $3$ & -- & $53.19$ \smallpm{1.47} & -- & $n$ & -- & $81.36$ \smallpm{0.70} & -- \\ \midrule
    \multirow{2}{*} {\textbf{RadSAM}} & $3$ & -- &  $\mathbf{84.99}$ \smallpm{0.74} & $\mathbf{84.05}$ \smallpm{0.86} & $n$ & -- & $\mathbf{91.08}$ \smallpm{0.35} &  $\mathbf{85.09}$ \smallpm{0.70} \\ 
    & $3$ & 20 & $\underline{91.11}$ \smallpm{0.39}  & $\underline{90.63}$ \smallpm{0.46} & $n$ & $10n$ & $\underline{96.73}$ \smallpm{0.13}  &  $\underline{96.78}$ \smallpm{0.13}\\
    \bottomrule
\end{tabular}
\caption{Dice scores on AMOS for RadSAM, SAM, MedSAM, SAM-Med3D and nnU-Net.
MedSAM does not support point prompting and edition, and nnU-Net does not support prompting.
$n$ represents the number of slices. For volume-level prompts, the $3$ represents the initial prompt and the two boundaries. We display the 95\% confidence intervals for all results we evaluate. The best results without edition and with edition are respectively in bold and underlined.}
\label{tab:main_results}
\end{table*}

\subsection{Edition in iterative segmentation}
Allowing editing through the iterative pipeline is more challenging than with a 2D mode, but it is important as errors can accumulate over the steps. We propose a method to integrate them to perform editions of the whole 3D mask with a single edition point.
These correction points are analogous to those used in 2D scenarios and are placed where the mask is either incorrectly present or absent. They are sampled uniformly among the errors.

The propagation strategy is similar to that used for the first segmentation. When the user adds a correction point to the slice $i$, we feed the previous mask $m_i$ along with the correction point to obtain a corrected mask $m'_i$. We then start the propagation again with this new corrected mask. 
If the model encounters a slice with previous correction points, we can re-use them in addition to the previous mask to give the model more information.

%% file: sec/3_results.tex
\section{Results}
\label{sec:results}

\subsection{Main Results}

We compare RadSAM with two publicly available 2D prompted segmentation models: MedSAM~\cite{ma_segment_2024}, trained on multiple medical datasets including AMOS~\cite{ji_amos_2022}, and SAM~\cite{kirillov_segment_2023}, trained on large datasets of natural images.
Unless specified otherwise, we evaluate all models on the AMOS validation set, composed of 100  CT volumes. All evaluations report the 3D Dice metric, defined as follows:

$$\text{Dice}_\text{score} = \frac{2 \times M \cap G}{M + G}$$
where $M$ is the mask produced by the model and $G$ is the ground truth. We compute this score on the whole organs for all benchmarks, even in the slice-level prompting evaluations.

As explained in Section~\ref{sec:method}, RadSAM can be used in different modes: similar to MedSAM, a volume-level prompting with one prompt for the whole volume or a slice-level prompting with a 2D prompt per slice. For a fair evaluation, we evaluate our RadSAM and MedSAM using both approaches: volume-level prompting and slice-level prompting.

\paragraph{Volume-Level Prompting}
In Table~\ref{tab:main_results} we display the Dice scores on AMOS for multiple models. As a baseline, we show a 3D nnU-Net~\cite{isensee_nnu-net_2021}, a semantic segmentation model that does not use prompting. We also show SAM and MedSAM, which are 2D-based models with bounding boxes for iterative prompting, as explained in Section~\ref{sec:method-ar:iterative}. Finally, we show SAM-Med3D a 3D model.
We first evaluate models with volume-level prompting, with slice boundaries on the top and bottom of the organ, which stops the iterative process. 

RadSAM reaches a dice of 84.99 with a single volume-level bounding box prompt, largely beating MedSAM, which gets a dice of 53.19. With a point prompt, which can be much more ambiguous, our model obtains 84.05, losing only 0.94 points compared to the box prompt. With 20 edition points, RadSAM gains 6.12 points, beating the semantic segmentation model nnU-Net. SAM and MedSAM perform very poorly on this setup. This can be partially attributed to the less precise bounding box prompt used in the iterative segmentation process. RadSAM also shows a gain of 4,11 points over SAM-Med3D using only one point and a score of 88.61 using a point prompt and 9 edition points, yielding a 4,62 points improvement over SAM-Med3D.

\begin{figure}[t]
    \centering
    \includegraphics[width=\linewidth]{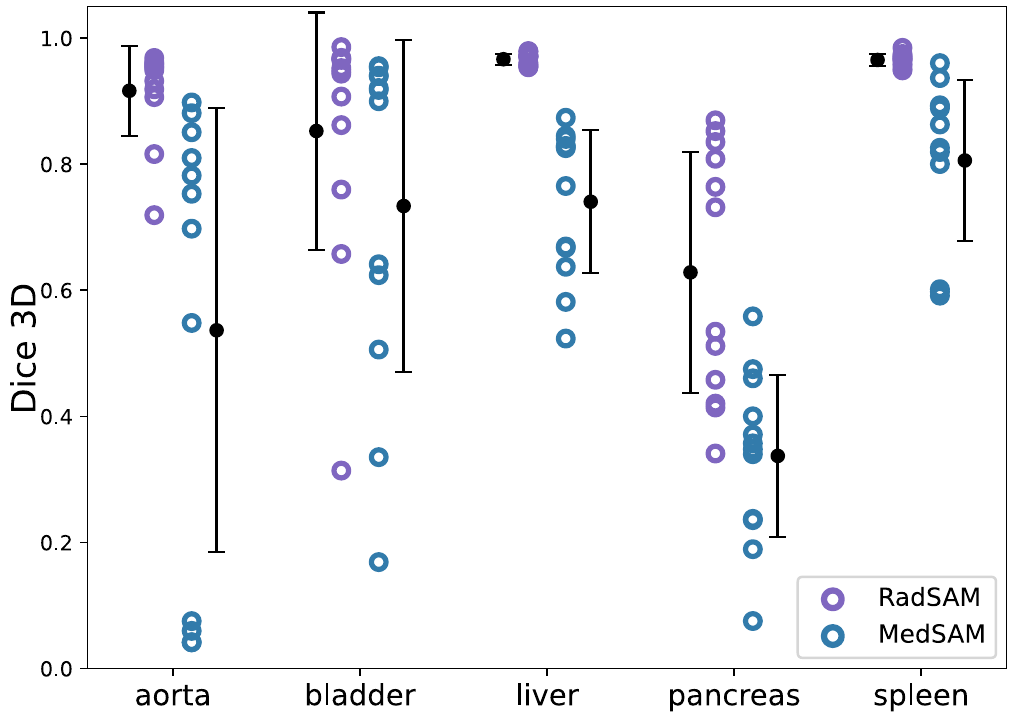}
    \caption{Quantitative evaluation results using the iterative pipeline on some classes of AMOS. We represented the standard deviation on all categories.}
    \label{result-auto-regressif-quali}
\end{figure}

Figure~\ref{result-auto-regressif-quali} shows organ-level scores along with their variances. RadSAM provides more accurate 3D masks and a much smaller variance interval on most anatomical structures.
We show additional details about the results in the supplementary materials.

\paragraph{Slice-Level Prompting}
\label{sec:eval-2D}
As mentioned before, we perform volume-level prompting for SAM and MedSAM with a bounding box for the iterative segmentation. This leads to suboptimal results, as a bounding box is much less precise than a mask.
Therefore, to fairly compare RadSAM with those 2D models, we also compare them in Table~\ref{tab:main_results} to MedSAM's original setup: slice-level prompting. For an object with $n$ slices, we give $n$ initial prompts. The dice scores are still computed on the final 3D object. 
Additionally, we assessed the model's performance with 10 editing points added to each slice. RadSAM obtains a dice score of 91.08, beating the scores of nnU-Net and MedSAM by a significant margin.
The difference between RadSAM and MedSAM scores can be explained by their different data augmentation strategies, windowing, and more specialized training.

\paragraph{Qualitative examples}

\begin{figure}
    \centering
    \includegraphics[width=\linewidth]{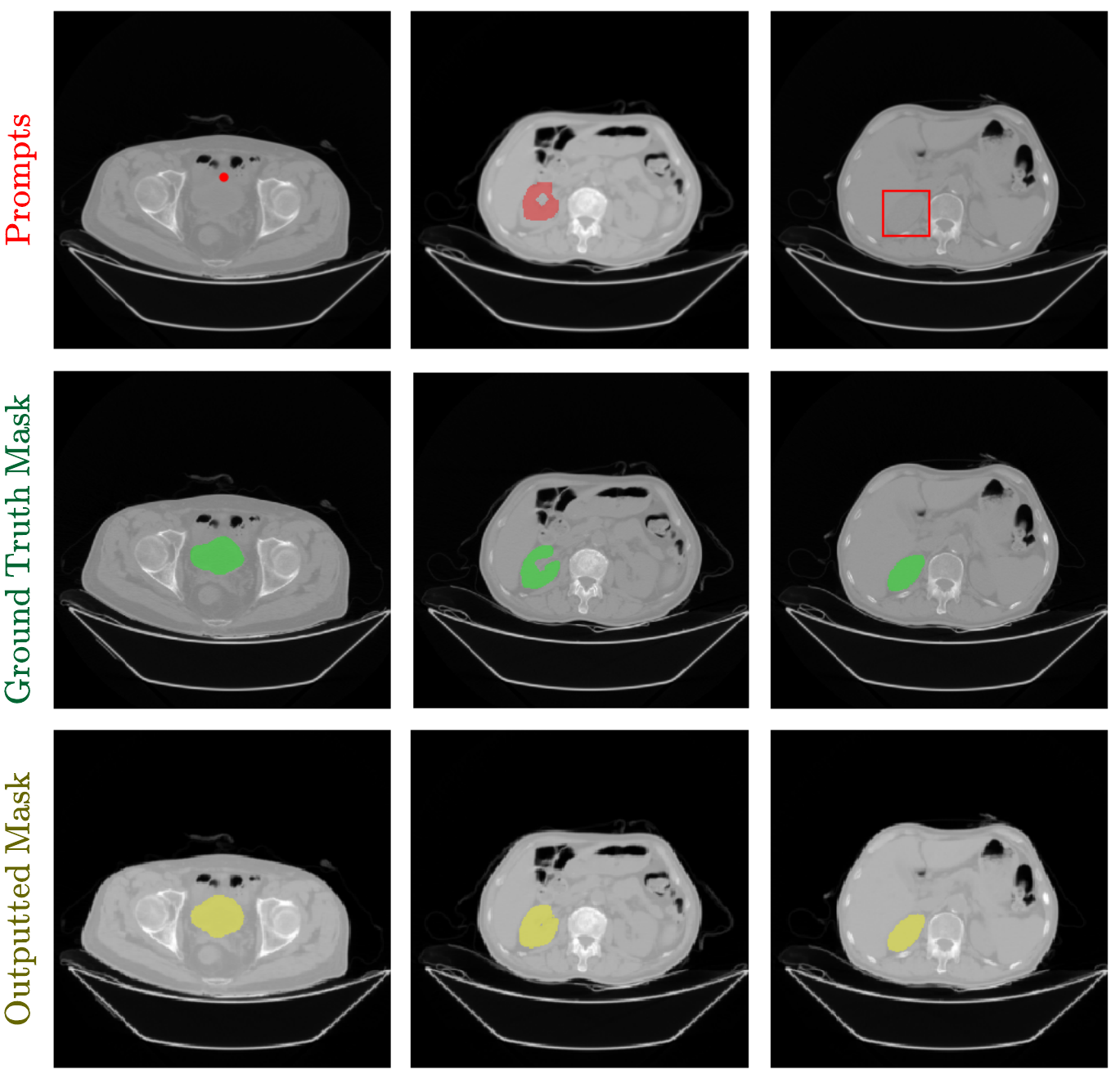}
    \caption{Qualitative results from various prompts. The prompt example appears in red, and on top, the ground truth mask appears in green under. At the bottom in yellow is the mask predicted by the model.}
    \label{fig:qualiresult}
\end{figure}

Figure~\ref{fig:qualiresult} shows a qualitative example of our model's predictions for each type of prompt: point, box, and mask. We display the prompt, the ground-truth mask, and the model's prediction. We show additional qualitative examples in the supplementary materials.


\begin{figure*}[t]
    \centering
    \begin{subfigure}[b]{0.44\textwidth}
        \centering
        \includegraphics[width=\linewidth]{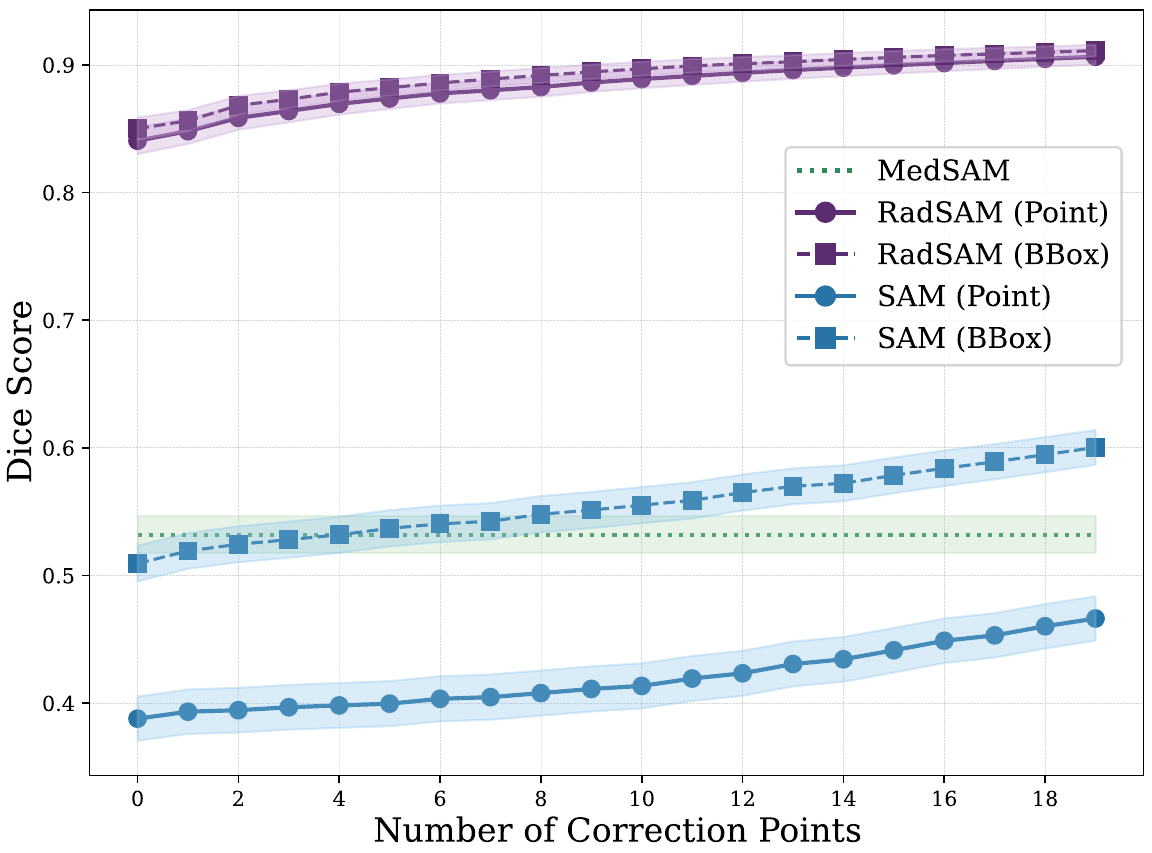}
        \caption{Volume-level prompting}
        \label{fig:result-amos-edition-volume}
    \end{subfigure}
    \hfill
     \begin{subfigure}[b]{0.44\textwidth}
        \centering
        \includegraphics[width=\linewidth]{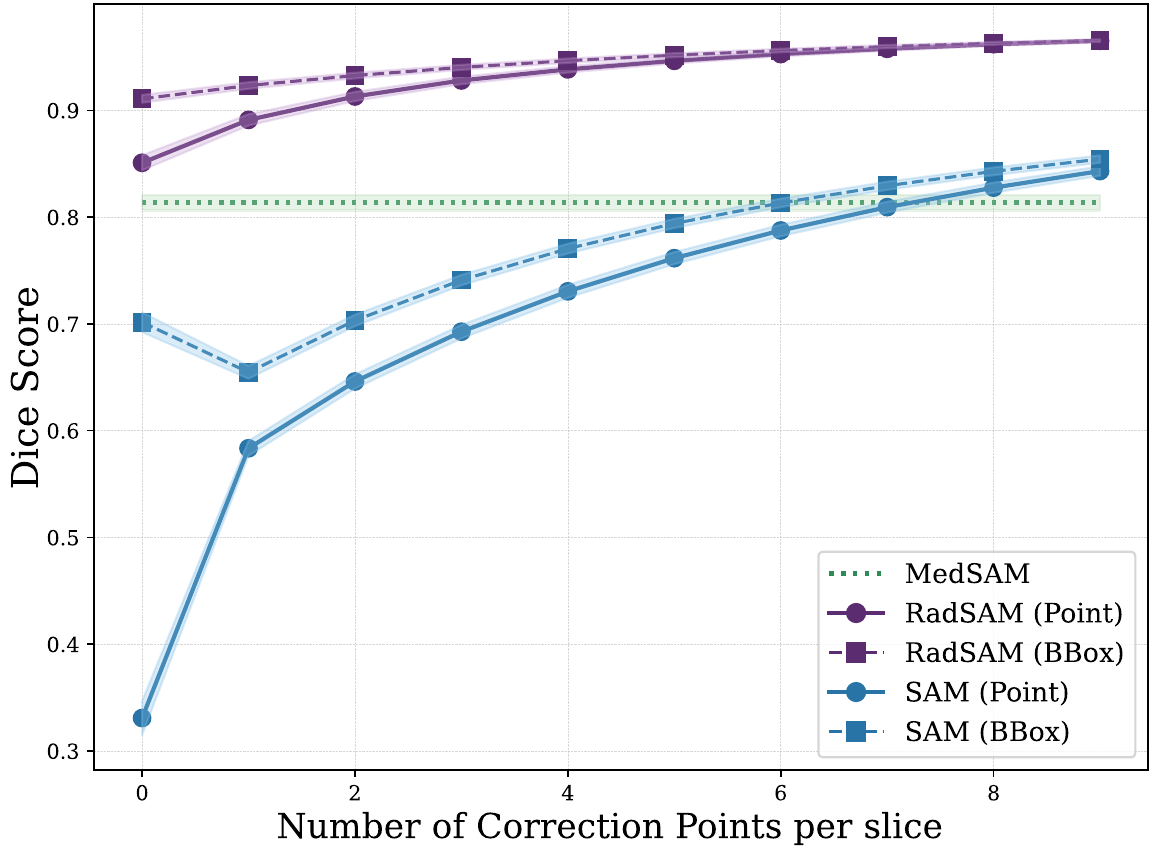}
        \caption{Slice-level prompting}
        \label{fig:result-amos-edition-slice}
         \end{subfigure}
    \caption{Dice scores with various numbers of edition points on AMOS. MedSAM is represented as a straight line, as it does not support edition.}
    \label{fig:result-amos-edition}
\end{figure*}

\subsection{Edition Capabilities}
We report the editing capabilities of the three models with volume-level and slice-level promptings in Figure~\ref{fig:result-amos-edition}.
We show important gains with editing: in volume-level prompting, the model gains 6,12 dice points when going from 0 to 20 edition points. In slice-level prompting, the model gains 5,65 points with 10 points per slice. We see that when given enough edition points, the difference in performances of the box-prompted and point-prompted models decreases significantly, reaching only 0,48 points and 0.03 points for volume-level and slice-level prompting. The additional points help to reduce the ambiguity caused by the point prompting.

Another interesting observation is that SAM, while not specifically trained on medical images, beats MedSAM on both setups with sufficient edition prompts. This highlights the importance of integrating the edition mode.

\subsection{Transfer Learning}
Medical models must be robust to images from various domains, as images coming from different hospitals or machines can have different distributions.
We use the TotalSegmentator dataset to assess our model's performance on out-of-domain images. We only consider bounding box prompts for all models, as they provide the best overall results. 
We split TotalSegmentator classes into two subsets: ``known'' classes present in AMOS and ``unknown'' classes, which are not. We show the results in Table~\ref{tab:results_transfer} and give more detailed results in the supplementary materials.
RadSAM demonstrates superior generalization, outperforming MedSAM on out-of-domain known classes: it obtains 60.98 dice (+16.52 with respect to MedSAM) with volume-level prompting and 78.72 (+11,46 points) with slice-level prompting. As expected, SAM obtains a lower score on those classes.

For unknown classes, we observe that while slice-level prompting gives average performances (62.94 dice, losing around 20 points to the known classes), the volume-level prompting model's performances degrade much more, losing around 30 points to reach 30.57. The error accumulates in the iterative segmentation process while prompting each slice independently, preventing the model from segmenting the wrong structure because of failure on a specific slice.  
Overall, the SAM model performs better or similarly to the other models on unknown classes. This suggests that models have not been trained on diverse enough data to gain generalization abilities to novel objects.

\begin{table}[h]
    \begin{subtable}[h]{\linewidth}
        \centering
        \begin{tabular}{@{}lll@{}}                 
        \toprule
        \multicolumn{1}{l}{}  & \multicolumn{1}{l}{Known classes}  & \multicolumn{1}{l}{Unknown classes}  \\ 
        \midrule
        SAM      & $29.71$ \smallpm{2.72}       & $33.86$ \smallpm{0.65}    \\
        MedSAM   & $44.46$ \smallpm{3.28}       & $24.85$ \smallpm{1.87}    \\ 
        RadSAM   & $\mathbf{60.98}$ \smallpm{3.22}    & $\mathbf{30.57}$ \smallpm{0.68}  \\
        \bottomrule
        \end{tabular}
        \caption{Volume-level prompting}
        \label{tab:results_transfer-3D}
    \end{subtable}
    \hfill
    \begin{subtable}[h]{\linewidth}
        \centering
       \begin{tabular}{@{}lll@{}}                 
            \toprule
            \multicolumn{1}{l}{}  &  \multicolumn{1}{l}{Known classes}   &  \multicolumn{1}{l}{Unknown classes} \\ 
            \midrule
            SAM     & $50.77$ \smallpm{2.30}           & $54.53$ \smallpm{0.46}   \\ 
            MedSAM  & $67.26$ \smallpm{2.08}           & $50.33$ \smallpm{0.47} \\
            RadSAM  & $\mathbf{78.72}$ \smallpm{1.74}  & $\mathbf{62.94}$ \smallpm{0.41} \\
            \bottomrule
            \end{tabular}
            \caption{Slice-level prompting}
            \label{tab:results_transfer-2D}
    \end{subtable}
     \caption{Generalization performances of a model trained on AMOS, evaluated on TotalSegmentator (TS). Known are the classes from TS that are present in AMOS, and unknown are all the other classes from TS.}
     \label{tab:results_transfer}
\end{table}



\subsection{Pre-training on larger datasets}
We evaluate whether pre-training our model on additional out-of-domain datasets can help increase performances on AMOS. Table~\ref{tab:results_pretraining_ts} shows that pre-training our model on TotalSegmentator before fine-tuning on AMOS outperforms the original model trained only on AMOS for most prompt types.

\begin{table*}[t]
\centering
\small
\begin{tabular}{llcccc}
\toprule
& & \multicolumn{2}{c}{Volume-level prompt} & \multicolumn{2}{c}{Slice-level prompt} \\
\cmidrule(lr){3-4} \cmidrule(lr){5-6}
\multicolumn{1}{l}{Model} & \small Training  & \small Bbox & \small Point & \small Bbox & \small Point \\ 
\midrule
MedSAM & 1.5M images including AMOS  & $53.19 \pm 1.47$& --   & $81.36 \pm 0.7$  & -- \\ 
RadSAM& TS*   & $66.56 \pm 1.38$& $59.54 \pm 1.66$  & $82.51 \pm 0.8$    & $59.38 \pm 1.5$    \\ 
RadSAM& AMOS  & $\mathbf{84.99 \pm 0.74}$& $84.05 \pm 0.86$ & $91.08 \pm 0.35$  & $85.09 \pm 0.70$ \\
RadSAM& TS $\rightarrow$ AMOS** & $82.15 \pm 0.87$& $\mathbf{81.75 \pm 0.97}$ & $\mathbf{91.40 \pm 0.3}$  & $\mathbf{86.20 \pm 0.6}$ \\
\bottomrule 
\end{tabular}
\caption{3D Dice scores on AMOS using for various training datasets. 
*TS corresponds to RadSAM but with a pre-training on TotalSegmentator rather than AMOS.
**TS $\rightarrow$ AMOS corresponds to RadSAM but with a pre-training on TotalSegmentator and then a fine-tuning on AMOS.}
\label{tab:results_pretraining_ts}
\end{table*}

\section{Ablation Study}

We perform ablations of some critical parameters of our approach. The supplementary materials include additional experiments on prompting ablations and different datasets.

\vspace{-3mm}

\paragraph{Impact of iterative prompt in the volume-level pipeline}
We evaluate the choice of using the mask prompt for the iterative inference pipeline. We compare this prompt to using a bounding box around the mask of the previous slice. Our results, in Table~\ref{tab:ablation_autoregraps}, show that the mask prompt drastically increases the dice score, going from 66.56 to 84.99 (+18,43) with an initial bounding box prompt, with similar gains with the initial point prompt.
This finding highlights the importance of our mask-prompt training and partially explains why MedSAM obtains low results in volume-level prompt settings.

\vspace{-1mm}

\begin{table}[H]
\centering
\begin{tabular}{lllll}
\toprule
Iterative prompt & \multicolumn{2}{c}{Mask} & \multicolumn{2}{c}{Bbox} \\
\cmidrule(lr){2-3} \cmidrule(lr){4-5}
\small Initial prompt  & \small Bbox  & \small Point  & \small Bbox  & \small Point \\ \midrule
MedSAM  & \color{gray} 12.34        & \color{gray} 1.17           & 58.26& \color{gray} 3.22\\ 
RadSAM  & 84.99& 84.05  & 66.56& 59.54 \\ \bottomrule
\end{tabular}
\caption{Impact of the iterative prompt (mask or bbox) on the 3D dice. 
Results in grey show unsupported prompts.}
\label{tab:ablation_autoregraps}
\end{table}

\vspace{-6mm}

\paragraph{Oracle evaluation}
Table~\ref{tab:ablarion-oracle} compares the main predicted mask and the oracle mask (the best among the three outputted masks). The Oracle obtains significantly better performance at a low user cost: one additional click to select the desired mask.

\begin{table}[H]
\centering
\small
\begin{tabular}{lll}
\toprule
              & Bbox            & Point               \\ \midrule
Dice 3D       & 91.08           & 85.09              \\ 
Dice oracle   & 92.52           & 90.25             \\ \bottomrule
\end{tabular}
\caption{Oracle evaluation of RadSAM}
\label{tab:ablarion-oracle}
\end{table}

       

\paragraph{Scaling the model size}
All our previous experiments used the ViT-B architecture to reduce computing costs. We evaluate the performance of ViT-L in Table~\ref{tab:vit-comp} and show that scaling significantly increases performance: the model gains around 0.5 to 1 dice point for each prompt type.

\begin{table}[h]
\centering
\begin{tabular}{lll}
\toprule
\multicolumn{1}{c}{\textbf{}} & Bbox & Point \\ 
\midrule
ViT-B             & 91.08        & 85.09 \\ 
ViT-L             & 91.59        & 86.15 \\ 
\bottomrule
\end{tabular}
\caption{Dice scores for RadSAM with ViT-B and ViT-L backbones on AMOS using slice-level prompting}
\label{tab:vit-comp}
\end{table}
\vspace{-5mm}

\paragraph{Varying training datasets}
We compare the performance of our model under three different training scenarios in Table~\ref{tab:ablation:train_set}: (1) training only on TotalSegmentator, (2) training on TotalSegmentator followed by fine-tuning on AMOS, and (3) joint training on both datasets. The results reveal that fine-tuning on AMOS significantly boosts performance. However, this improvement comes at the cost of reduced performance on TotalSegmentor. In contrast, training on both datasets simultaneously achieves high performance, approaching the best performance achieved in each dataset. 

\begin{table}[h]
\centering
\begin{tabular}{lllll}
\toprule
 & \multicolumn{2}{c}{TS}     & \multicolumn{2}{c}{AMOS}    \\ 
 \cmidrule(lr){2-3} \cmidrule(lr){4-5}
 \small Training strategy & \small Bbox & \small Point  & \small Bbox   & \small Point \\ \midrule
TS & $90.87$& $88.78$  & $82.51$  & $59.38$      \\
TS $\rightarrow$ AMOS & $76.65$  & $62.47$  & $91.40$  & $86.20$\\
TS + AMOS  & $90.20$  & $87.67$  & $90.05$ & $83.27$\\
\bottomrule
\end{tabular}
\caption{Comparing training dataset of RadSAM, with evaluations on TS (all classes) and AMOS. Experiments use slice-level prompting.}
\label{tab:ablation:train_set}
\end{table}
\vspace{-5mm}

\paragraph{Removing key components}
Finally, we evaluate the impact of adding the mask-prompting and edition on the slice-level segmentation performance without edition, i.e., when those features are not used.
Table~\ref{tab:ablation_mask_edit} shows that these features do not improve nor harm the model's raw 2D performance while enabling new capabilities, such as improving the iterative segmentation, as shown previously.

\begin{table}[h]
\centering
\begin{tabular}{ccll}
\toprule
With mask     & With edition     & Bbox   & Point   \\ 
\midrule
\ding{51}     &\ding{51} & $91.08$& $85.09$ \\
\ding{55}     &\ding{51} & $90.74$& $84.23$ \\
\ding{51}     &\ding{55} & $91.02$& $83.64$ \\
\ding{55}     &\ding{55} & $90.96$& $85.35$  \\ 
\bottomrule
\end{tabular}
\caption{3D Dice scores of RadSAM with slice-level prompting on AMOS when removing the mask or the edit during the training}
\label{tab:ablation_mask_edit}
\end{table}

\section{Conclusion}
We propose a simple method for 3D prompted segmentation using volume-level prompts from a 2D segmentation model. Naive approaches with existing models, like using bounding boxes to forward the prompt from slice to slice, perform very poorly. However, adding a novel prompt type, the mask, as the iterative prompt drastically improves those results. RadSAM demonstrates performances close to the state-of-the-art segmentation model nnU-Net, surpassing it with edition points. Additionally, we show that RadSAM performs well when evaluated on other datasets with the same classes.
This work lays the foundation for more effective clinical utilization of segmentation models. Their interactivity through prompting and editing is essential to give users control over the output mask for most medical tasks where decisions impact clinical care.
Given their potential utility in real-world applications, developing their zero-shot capabilities is particularly important. Future work is needed to match the impressive performances on novel objects, obtained by generalist segmentation models such as SAM on natural images.

\section{Acknowledgments}
This work was granted access to the HPC resources of IDRIS under the allocation 2024-AD011013489R2 made by GENCI.


%% file: sec/X_suppl.tex
\clearpage
\setcounter{page}{1}
\maketitlesupplementary


\section{Qualitative results}

\begin{figure}[H]
    \centering
    \includegraphics[width=0.85\linewidth]{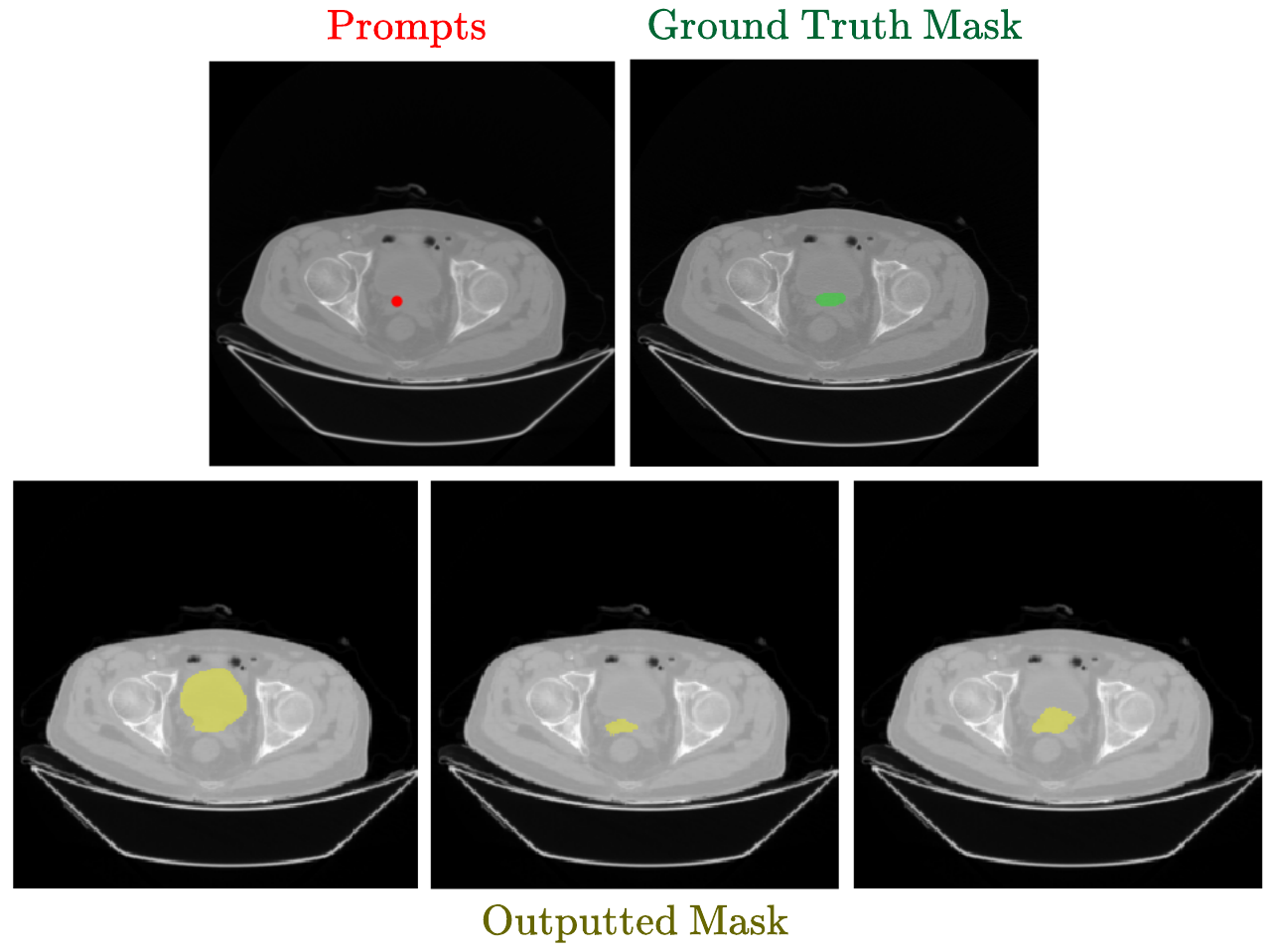}
    \caption{Example of the 3 different masks outputted by the model using a point prompt. The prompt example appears in red, the ground truth mask appears in green. At the bottom in yellow is the mask predicted by the model.}
    \label{result-outputs}
\end{figure}

\begin{figure*}[t]
    \centering
    \includegraphics[width=\linewidth]{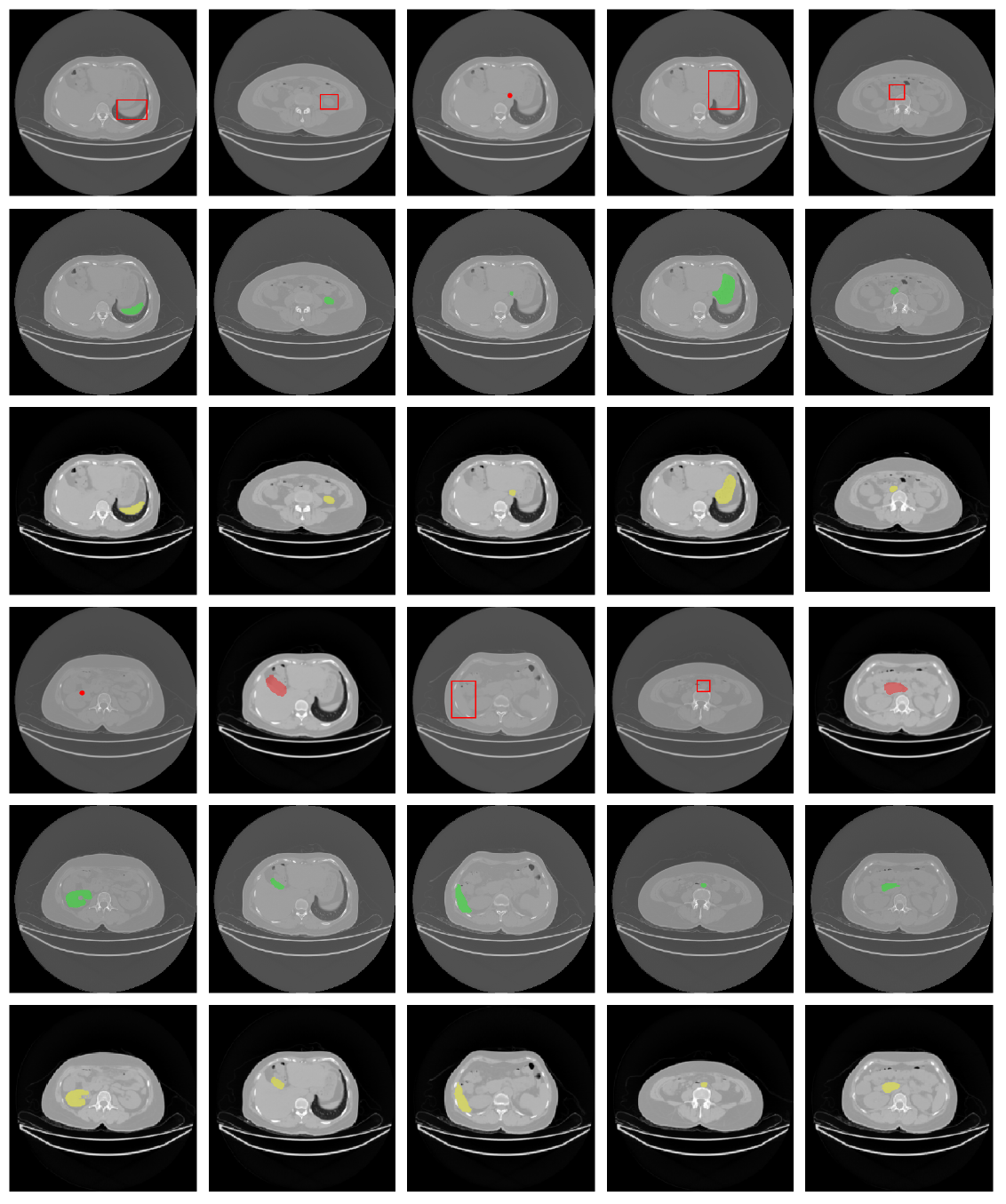}
    \caption{Qualitative results of the model. The prompt example appears in red, the ground truth mask appears in green. At the bottom in yellow is the mask predicted by the model.}
    \label{result-qualit-outputs}
\end{figure*}

\paragraph{Multi-mask prediction} We present in Figure \ref{result-outputs} an example of the 3 masks predicted by RadSAM when given a point. The given point is ambiguous: at the border of multiple anatomical structures. Thus, the model proposes to segment multiple structures.  
\paragraph{Additional examples} We show more qualitative results in Figure \ref{result-qualit-outputs} using point, bbox, and mask as prompts.


\section{Result by categories}
We group the results on AMOS, comparing RadSAM and MedSAM. RadSAM has better results than MedSAM using a volume-level prompt or slice-level prompt, especially in the volume-level prompt, where errors can accumulate and give a difference of 61.0 points for 3D dice.


    



\section{Transfer results detailed}

In Table \ref{tab:results-transfer-detailed}, we detail the results from Section~5.3 on transfer from AMOS to TotalSegmentator. Using a slice-level prompt on unknown classes, RadSAM goes from a dice of 78.71 to 92.14, leveraging its edition capabilities, showing a difference of 13.42 points. It highlights the ability of RadSAM to assist in the segmentation of unknown structures and is the first step toward zero-shot models.

\begin{figure}[H]
    \centering
    \begin{subfigure}[b]{0.45\textwidth}
    \includegraphics[width=\linewidth]{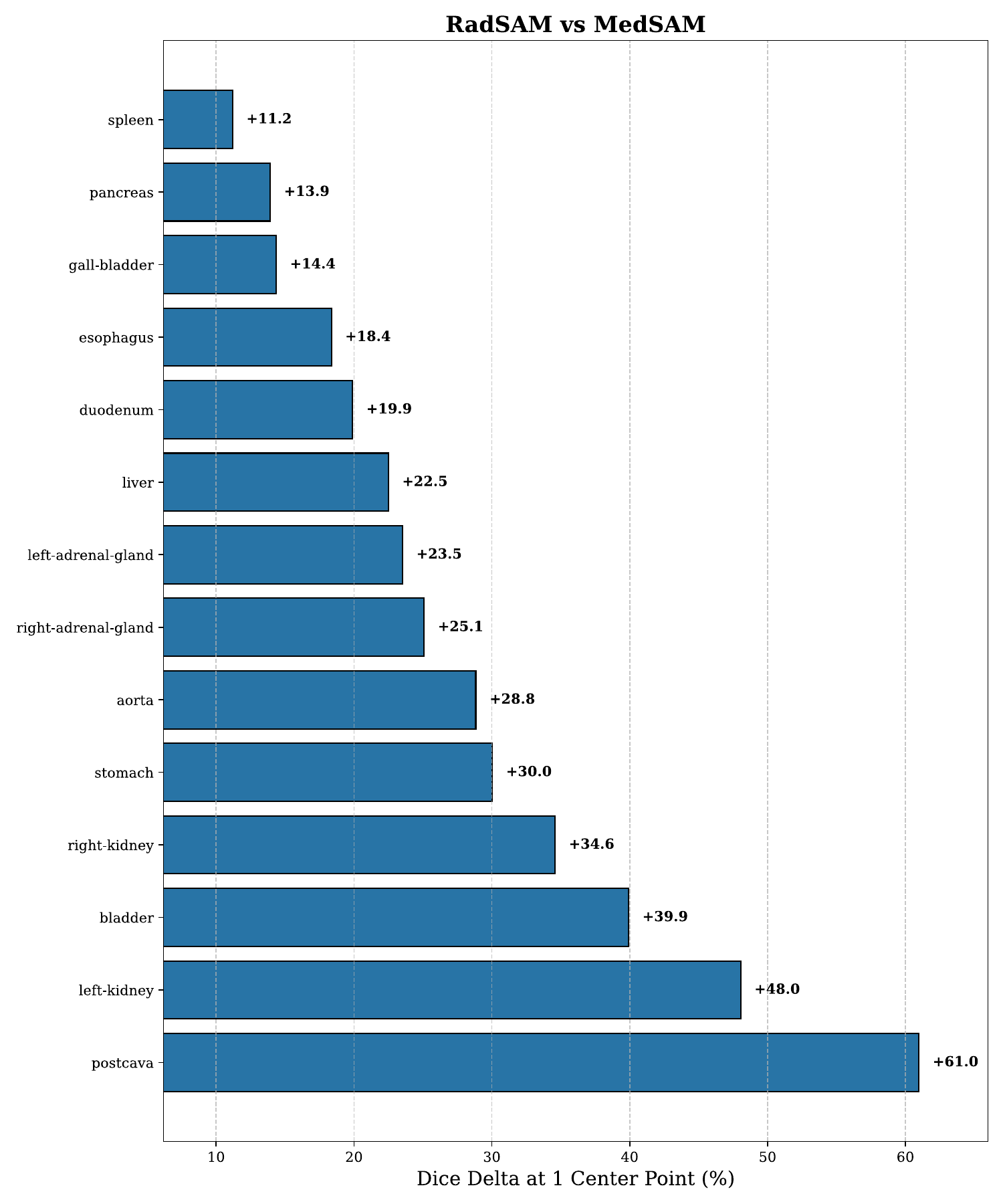}
    \caption{Volume-level}
    \end{subfigure}
    \hfill
    \begin{subfigure}[b]{0.45\textwidth}
    \includegraphics[width=\linewidth]{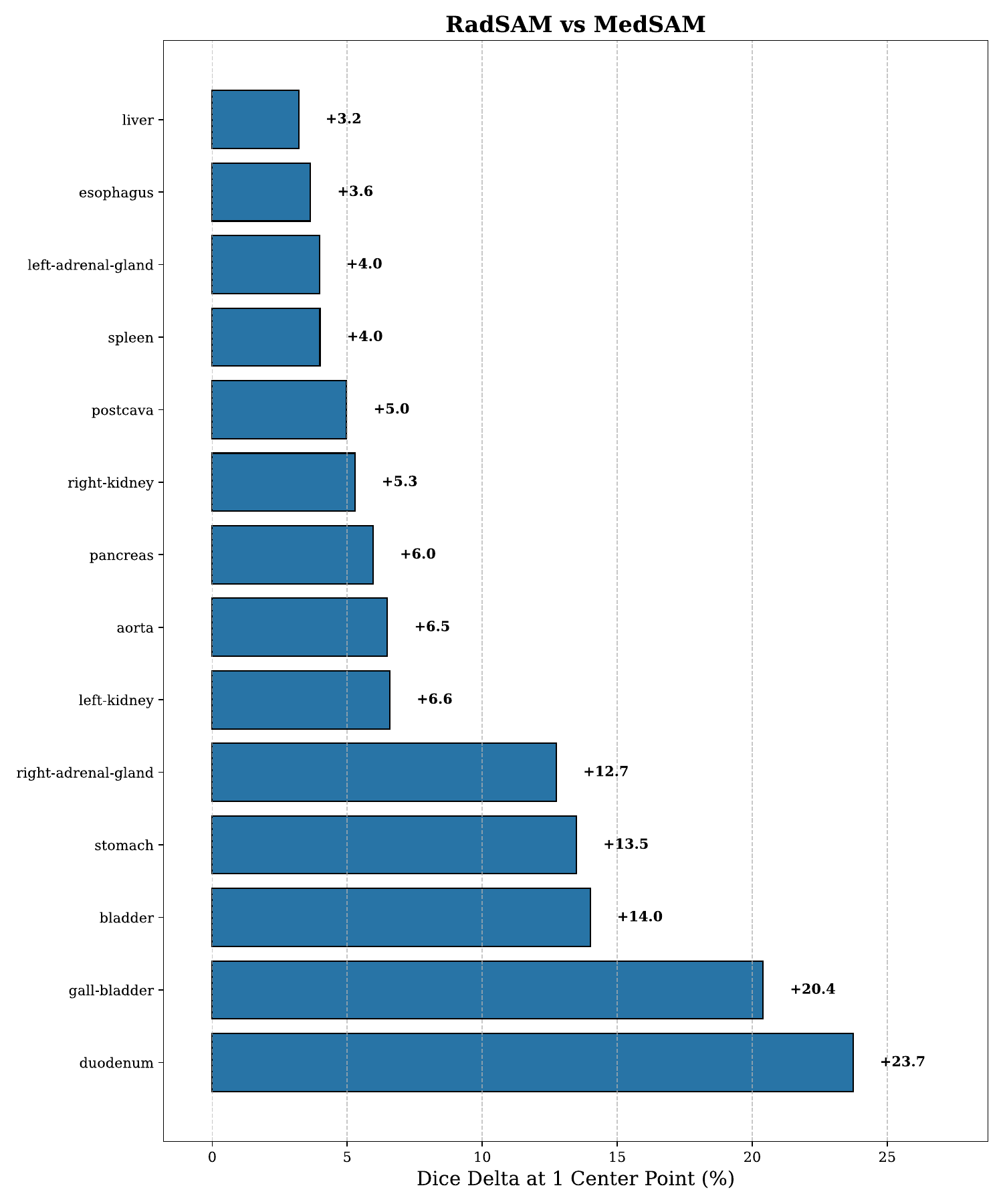}
    \caption{Slice-level}
    \end{subfigure}
    \caption{Bar plot of the different classes, showing the difference between RadSAM and MedSAM.}
    \label{result-histogram-supp}
\end{figure}

\begin{table*}[h!]
\centering
\footnotesize
\begin{tabular}{llcrcccccc}
    \toprule
    & \footnotesize Class type &\multicolumn{4}{c}{\footnotesize Volume-level prompt} &  \multicolumn{4}{c}{\footnotesize Slice-level prompt}  \\
    \cmidrule(r){3-6} \cmidrule(l){7-10}
    \footnotesize{Model} & &\footnotesize{\#Prompts} & \footnotesize{\#Edits} & {\footnotesize Bbox} & \footnotesize Point & 
    \footnotesize{\#Prompts} & \footnotesize{\#Edits} &{\footnotesize Bbox} & Point \\
    \midrule
    SAM & Known & $3$ & --  & $29.71$ \smallpm{2.72} & $28.47$ \smallpm{3.20} & $n$  & -- & $50.78$ \smallpm{2.28} & $54.53$ \smallpm{0.46} \\ 
    MedSAM& Known &$3$ & -- & $44.46$ \smallpm{2.80} & -- & $n$ & -- & $67.26$ \smallpm{2.08} & -- \\ 
    \multirow{2}{*} {\textbf{RadSAM}} & Known &$3$ & -- &  $\mathbf{60.98}$ \smallpm{3.22} & $\mathbf{64.71}$ \smallpm{2.87} & $n$ & -- & $\mathbf{78.72}$ \smallpm{1.68} &  $\mathbf{65.36}$ \smallpm{2.56} \\ 
    & Known &$3$ & 4 & $\underline{66.91}$ \smallpm{3.13}  & $\underline{66.91}$ \smallpm{2.69} & $n$ & $10n$ & $\underline{92.14}$ \smallpm{0.74}  &  $\underline{91.65}$ \smallpm{1.02}\\ \midrule
    SAM & Unknown & $3$ & --  & $\mathbf{33.86}$ \smallpm{0.65} & $\mathbf{35.53}$ \smallpm{0.75} & $n$  & -- & $54.53$ \smallpm{0.46} & $36.73$ \smallpm{0.69} \\ 
    MedSAM& Unknown &$3$ & -- & $24.85$ \smallpm{0.53} & -- & $n$ & -- & $50.33$ \smallpm{0.47} & -- \\ 
    \multirow{2}{*} {\textbf{RadSAM}} & Unknown &$3$ & -- &  $30.57$ \smallpm{0.68} & $35.91$ \smallpm{0.66} & $n$ & -- & $\mathbf{62.98}$ \smallpm{0.41} &  $\mathbf{49.55}$ \smallpm{0.52} \\ 
    & Unknown &$3$ & 4 & $\underline{35.64}$ \smallpm{0.32}  & $\underline{38.50}$ \smallpm{0.67} & $n$ & $10n$ & $\underline{85.97}$ \smallpm{0.22}  &  $\underline{85.54}$ \smallpm{0.27}\\
    \bottomrule
\end{tabular}
\caption{Dice scores on TS for RadSAM, SAM, MedSAM
MedSAM does not support point prompting and edition.
Unknown correspond to classes that are present in AMOS, Known to those that are present.
$n$ represents the number of slices. For volume-level prompts, the $3$ represents the initial prompt and the two boundaries. We display the 95\% confidence intervals for all results we evaluate. The best results without edition and with edition are respectively in bold and underlined.}
\label{tab:results-transfer-detailed}
\end{table*}

\section{Results on Lesions}

Using the dataset Medical Segmentation Decathlon (MSD) \cite{antonelli2022medical}. We can see that our approach works to segment pathologies such as tumors. These results are detailed in Table \ref{tab:results-msd}, showcasing that RadSAM can segment Lung lesions with an average dice of 78.11 after a small finetuning.

\begin{table}[H]
\centering
\begin{tabular}{cc}
    \toprule
    Label & Dice 3D \\
    \midrule
    Lung lesion & 78.11 \\
    Liver lesion & 68.60 \\
    Colon lesion & 79.37 \\
    Pancreas lesion & 78.16 \\
    \bottomrule
\end{tabular}
\caption{Results on the MSD dataset after training on this dataset using bounding box as volume prompt and no edition.}
\label{tab:results-msd}
\end{table}